\makeatletter\def\graphicscache@inhibit{true}\makeatother
\documentclass[conference]{IEEEtran}

\usepackage[backend=bibtex,style=ieee,natbib=true,mincitenames=1,maxcitenames=1]{biblatex}
\usepackage{amsmath,amssymb,amsfonts}
\usepackage{algorithmic}
\usepackage{graphicx}
\usepackage{textcomp}
\usepackage{xcolor}
\usepackage[hidelinks]{hyperref}
\usepackage[capitalize]{cleveref}
\usepackage{booktabs}
\usepackage{placeins}
\usepackage{threeparttable}
\usepackage{adjustbox}
\usepackage{flushend}

\usepackage{siunitx}
\usepackage[draft,nomargin,inline]{fixme}
\fxsetup{theme=color}

\usepackage{tikz}
\usetikzlibrary{positioning}
\usetikzlibrary{fit}
\usetikzlibrary{calc}

\makeatletter

\tikzdeclarecoordinatesystem{rel}{%
	\tikzset{cs/.cd,x=0pt,y=0pt,#1}%
	\pgfpointlineattime{(\tikz@cs@x / 100)}%
	{\pgfpointanchor{\tikz@pp@name{\tikz@cs@node}}{south west}}%
	{\pgfpointanchor{\tikz@pp@name{\tikz@cs@node}}{south east}}%
	\edef\tikz@cs@x{\the\pgf@x}%
	\pgfpointlineattime{(\tikz@cs@y / 100)}%
	{\pgfpointanchor{\tikz@pp@name{\tikz@cs@node}}{south west}}%
	{\pgfpointanchor{\tikz@pp@name{\tikz@cs@node}}{north west}}%
	\pgfpoint{\tikz@cs@x}{\pgf@y}%
}

\makeatother

\IfFileExists{graphicscache.sty}{\usepackage{graphicscache}}

\begin{document}

\title{Predicting Physical Object Properties from Video\\
}

\author{\IEEEauthorblockN{Martin Link}
\IEEEauthorblockA{\textit{Autonomous Intelligent Systems} \\
\textit{University of Bonn}\\
Bonn, Germany \\
\texttt{mlink@uni-bonn.de}}
\and
\IEEEauthorblockN{Max Schwarz}
\IEEEauthorblockA{\textit{Autonomous Intelligent Systems} \\
\textit{University of Bonn}\\
Bonn, Germany \\
\texttt{schwarz@ais.uni-bonn.de}}
\and
\IEEEauthorblockN{Sven Behnke}
\IEEEauthorblockA{\textit{Autonomous Intelligent Systems} \\
\textit{University of Bonn}\\
Bonn, Germany \\
\texttt{behnke@cs.uni-bonn.de}}
}

\maketitle

\begin{tikzpicture}[remember picture,overlay]
\node[anchor=north,align=left,font=\sffamily,yshift=-0.2cm] at (current page.north) {%
  In: International Joint Conference on Neural Networks (IJCNN) 2022
};
\end{tikzpicture}%

\begin{abstract}
  We present a novel approach to estimating physical properties of objects from video.
  Our approach consists of a physics engine and a correction estimator.
  Starting from the initial observed state, object behavior is simulated forward in time.
  Based on the simulated and observed behavior, the correction estimator then determines refined
  physical parameters for each object. The method can be iterated for increased precision.
  Our approach is generic, as it allows for the use of an arbitrary---not necessarily differentiable---physics engine and correction estimator. For the latter, we evaluate both gradient-free hyperparameter optimization and a deep convolutional neural network.
  We demonstrate faster and more robust convergence of the learned method in several simulated 2D scenarios focusing on bin situations.
\end{abstract}


\section{Introduction}

Many tasks in robotics and autonomous systems require a reliable model of the world that is surrounding the agent.  Notably, this includes a model of the physical properties of foreign objects: Mass, surface friction, elasticity, moment-of-inertia, and density distribution/center-of-mass can all play a crucial role and---when guessed incorrectly---lead to failure cases.  One prominent example application is bin picking~\citep{morrison2018cartman,schwarz2018}, where a robotic agent has to detect and manipulate objects.

\begin{figure}[t]
\centering
\includegraphics[scale=1.0]{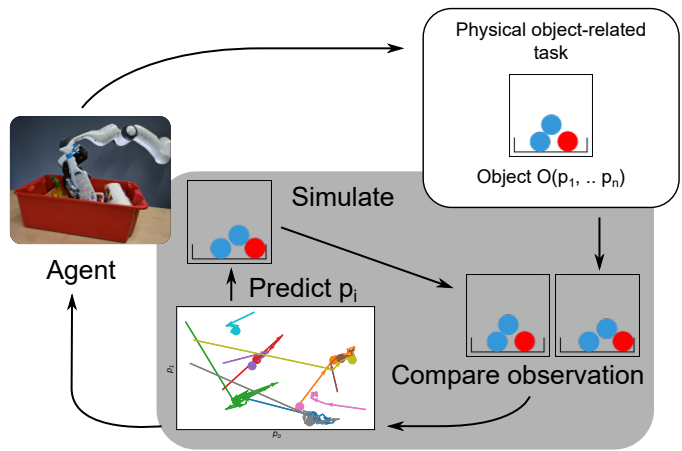}
\caption{Overview. An agent needs to complete a general task related to a physical object $O$ that is described by parameters $p_i$.  Our approach (grey box) allows the agent to learn these parameters by observing the behavior of the objects and comparing it to a simulated scene that uses the current best estimation of $p_i$. Comparing both observation and simulation yields a refined set of parameters that can either be used to attempt completing the task or re-iterated with the current observation.}
\label{intro_fig}
\end{figure}

We explore ways to improve the knowledge of physical parameters of objects, as illustrated in~\cref{intro_fig}.
One way to achieve this is the use of a physics engine which in itself is differentiable and therefore allows for backpropagation of observed errors between simulation and reality \textit{through} the engine, in order to update the physical parameters of a simulation. In this work, we investigate a more flexible approach: We propose a framework that uses a correction estimator---e.g. a neural network---to refine the physical parameters in a simulation iteratively, by comparing an observation to the simulation. Since we treat the physics engine as a black box, we remove the need for differentiability and instead rely on well-understood neural network and machine learning techniques.  This simplification allows for the use of commercially available physics engines that are already optimized for speed and efficiency but are not necessarily differentiable.

Our contributions include 1) a general iterative framework for optimization of
physical object parameters and 2) a thorough evaluation of learning-based and task-agnostic approaches
for the correction estimator.

\begin{figure*}[ht]
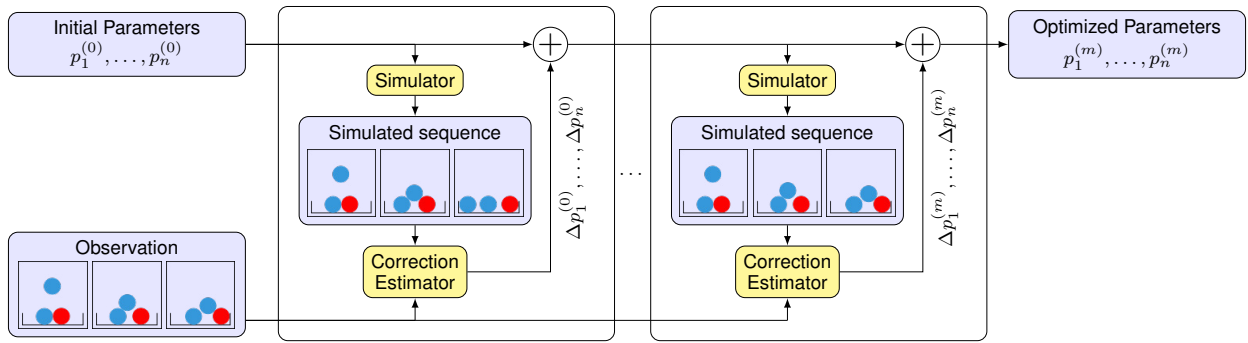

\centering
\adjustbox{scale=0.9}{%
\begin{tikzpicture}[font=\sffamily\footnotesize,
  m/.style={draw=black, rounded corners, fill=yellow!50, align=center},
  v/.style={m,fill=blue!10}
  ]

 \node[v,anchor=east,minimum width=3.5cm] (initial) at (-2.5, 1) {Initial Parameters\\[.1cm]
 $p_1^{(0)}, \dots, p_n^{(0)}$};
 \node[v,anchor=east,minimum width=3.5cm] (observed) at (-2.5, -2.55) {
    Observation\\[.1cm]
    
    \includegraphics[height=1cm]{figures/start_scene.png}
    \includegraphics[height=1cm]{figures/scene_2_1.png}
    \includegraphics[height=1cm]{figures/scene_2_2.png}
  };
 \coordinate (obsport) at ($(observed.south east)+(0,0.25)$);

 \coordinate (stepinput) at (0,1);
 \node[draw=black,circle,font=\Large,inner sep=0] (stepoutput) at (2.0,1) {$+$};
 \draw (initial) -- (stepinput);
 \draw[-latex] (initial) -- (stepoutput);
 
 \node[m,anchor=north] (sim) at (0,0.7) {Simulator};
 \draw[-latex] (stepinput) -- (sim);
 
 \node[v,below=0.3cm of sim] (simseq) {
    Simulated sequence\\[.1cm]

    \includegraphics[height=1cm]{figures/start_scene.png}
    \includegraphics[height=1cm]{figures/scene_1_1.png}
    \includegraphics[height=1cm]{figures/scene_1_2.png}
 };
 \draw[-latex] (sim) -- (simseq);
 
 \node[m,below=0.3cm of simseq] (pred) {Correction\\Estimator};
 \draw[-latex] (simseq) -- (pred);
 \draw[-latex] (pred.east) -| (stepoutput) node [pos=0.75,below,sloped] (delta) {$\Delta p_1^{(0)}, \dots, \Delta p_n^{(0)}$};
 
 \coordinate (obsinput) at (stepinput|-obsport);
 \draw (obsport) -- (obsinput);
 \draw[-latex] (obsinput) -- (pred);
 
 \node[fit=(sim)(stepoutput)(simseq)(pred)(delta)(obsinput), draw, rounded corners, inner sep=0.3cm] (step1) {};

 \begin{scope}[shift={(5.5,0)}]
 \coordinate (stepinput) at (0,1);
 \draw (stepoutput) -- (stepinput);
 \node[draw=black,circle,font=\Large,inner sep=0] (stepoutput) at (2.0,1) {$+$};
 \draw[-latex] (stepinput) -- (stepoutput);
 
 \node[m,anchor=north] (sim) at (0,0.7) {Simulator};
 \draw[-latex] (stepinput) -- (sim);
 
 \node[v,below=0.3cm of sim] (simseq) {
    Simulated sequence\\[.1cm]

    \includegraphics[height=1cm]{figures/start_scene.png}
    \includegraphics[height=1cm]{figures/scene_2_1.png}
    \includegraphics[height=1cm]{figures/scene_2_2.png}
 };
 \draw[-latex] (sim) -- (simseq);
 
 \node[m,below=0.3cm of simseq] (pred) {Correction\\Estimator};
 \draw[-latex] (simseq) -- (pred);
 \draw[-latex] (pred.east) -| (stepoutput) node [pos=0.75,below,sloped] (delta) {$\Delta p_1^{(m)}, \dots, \Delta p_n^{(m)}$};
 
 \coordinate (obsinput) at (stepinput|-obsport);
 \draw (obsport) -- (obsinput);
 \draw[-latex] (obsinput) -- (pred);
 
 \node[fit=(sim)(stepoutput)(simseq)(pred)(delta)(obsinput), draw, rounded corners, inner sep=0.3cm] (step2) {};
 \end{scope}
 
 \node at ($(step1)!0.5!(step2)$) {$\dots$};
 \node[v,right=1cm of stepoutput,minimum width=3.5cm] (optout) {Optimized Parameters\\[.1cm]
 $p_1^{(m)}, \dots, p_n^{(m)}$};
 \draw[-latex] (stepoutput) -- (optout);
 
\end{tikzpicture}%
}

\caption{Overview of our approach. The process starts with an initial
guess of physical parameters $p_1^{(0)},\dots,p_n^{(0)}$ and the observed
video sequence. In each iteration, the scene is simulated
using the current parameter estimate. The (learned) correction estimator then
produces a parameter update, which results in a better parameter estimate for
the next iteration. Figure inspired by \citep{li2020}.
}
\label{fig:overview}
\end{figure*}

\section{Related Work}

The task of predicting physical parameters of objects has been investigated before, for example from short video sequences in an unsupervised fashion~\citep{wu2016}. Often, properties are also predicted from single pictures, for example via micro-CT pictures for porous media \citep{alqahtani2018}, pictures of stacked crops~\citep{wani2019}, or pictures of liquid crystals~\citep{sigaki2020}.
In a more robotics-centric context, neural networks have been used to predict the hardness of objects with a GelSight sensor~\citep{yuan2017} and to identify parameters of unmanned aerial vehicles~\citep{ayyad2020}. In contrast to these earlier works, our dynamic approach utilizes iteration over observed scenes, which allows for constant refinement of the parameters by taking into account new information.

One application of our technique is bin picking.
Here, the success of planned operations is critically dependent on how well a robotic agent can be controlled. Often, controllers are optimized in simulated environments with reinforcement learning, particle swarms, or genetic algorithms, completely free of any derivatives \citep{aguilar2017,rubio2020,meda2018}.
As a consequence, the robot is often treated as black box, which prevents the use of efficient gradient-based deep learning methods.
Physics engines like MuJoCo~\citep{todorov2012} allow for the evaluation of gradients between states and actions through a robot, however, they do not allow backpropagation to the initial model parameters. More recent work investigated special physics engines that are end-to-end differentiable~\citep{degr2018,deav2018}, however, when testing the use of such engines, we found that they were much slower and less adaptable to new scenarios than commercial engines like NVidia PhysX\footnote{\url{https://developer.nvidia.com/physx-sdk}}.
Stillleben is a PhysX-based framework for the creation of simulated data to train agents in bin picking scenarios~\citep{schw2020}.
Recently, \citet{werling2021fast} introduced nimblephysics, a differentiable fork of the Dart physics engine. While nimblephysics allows backpropagation of object parameters,
this is currently limited to the object mass---highlighting the versatility of
our approach, as we enable optimization of any parameter as long as it influences the simulation
result in a learnable manner.

In this work, we present a framework that allows efficient estimation
of physical properties in an online fashion.
This includes keeping our framework efficient in order to be able to actively improve the scene knowledge at runtime. To this end, we build on an iterative approach that is used to estimate the 6D pose of objects~\citep{li2020}. There, a simulated pose is compared to an observed object in order to determine changes in the pose. This is done iteratively, in order to refine the predictions and allow for incremental improvements.
In our case, we \textit{observe} time series of frames to allow for efficient interpretation of the shown dynamics.
To estimate the relevant parameters, we compare the observed scene to a simulated scene. The comparison of both then allows to determine corrections to parameters of the simulation that correspond to physical properties of the involved objects. The neural network used as a correction estimator is based on the ResNet architecture~\citep{he2015}, but modified for the use of 3D convolutions to account for the time dimension.

\section{Method}

Our approach to predicting the physical parameters of objects
is inspired by earlier work on iterative prediction of pose estimation \cite{li2020}. In summary, we observe the scene as a sequence of video frames
and try to estimate the physical properties by producing a simulated scene with guessed parameters.
Importantly, this approach separates the simulation from the task of predicting the input parameters. In comparison to approaches where the simulation itself has to be differentiable, we can independently choose which physics engine we use for simulation and how to optimize the parameters.
This principle is illustrated in \cref{fig:overview}.
Starting from an initial guess of the parameters $p_1^{(0)}, \dots, p_n^{(0)}$,
we simulate the scene.
Note that we assume some sort of pose estimation that gives us the capability
to match the observable system state, so that we can start the simulation
in the same configuration as seen in the first observation frame.

Both sequences of pictures, observed and simulated, are then passed to the correction estimator, which outputs a set of parameter changes $\Delta p_i^{(t)}$ for all $n$ parameters entering the simulator. These parameter changes are then added to the initial parameters to yield the updated simulation parameters.

\begin{figure*}
\centering
\begin{tikzpicture}[font=\sffamily\scriptsize]
\node[inner sep=0] (pic) {\includegraphics[width=0.7\linewidth]{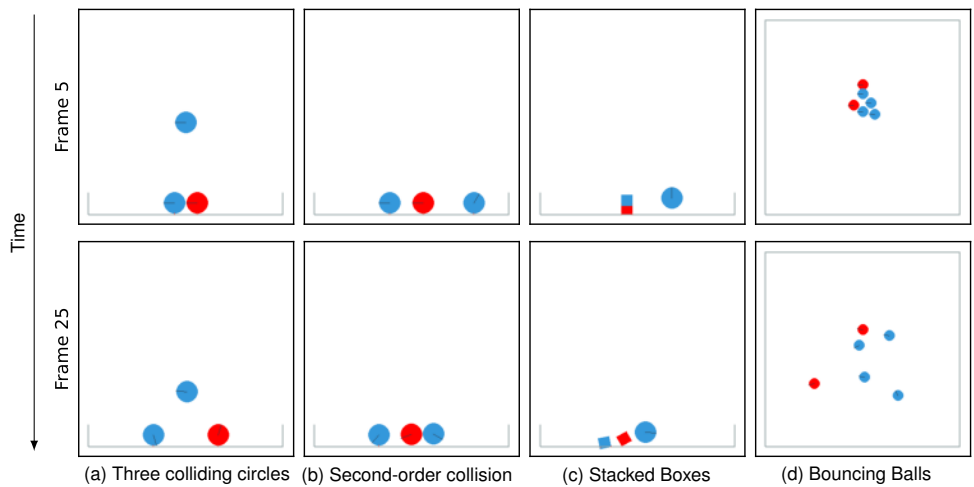}};
\draw[latex-] ($(pic.south west)+(-0.0,0.4)$) -- ($(pic.north west)+(-0.0,-0.4)$) node[midway,above,sloped] {Time};
\node[anchor=north,inner sep=0] at (rel cs:x=16,y=3,name=pic) {(a) Three colliding circles};
\node[anchor=north,inner sep=0] at (rel cs:x=39,y=3,name=pic) {(b) Second-order collision};
\node[anchor=north,inner sep=0] at (rel cs:x=63,y=3,name=pic) {(c) Stacked Boxes};
\node[anchor=north,inner sep=0] at (rel cs:x=86,y=3,name=pic) {(d) Bouncing Balls};
\end{tikzpicture}
\caption{Experimental scenes.}
\label{fig2}
\end{figure*}

The process can be iterated as often as needed: The updated parameters can be
used for the next simulation. In summary, each iteration consists of one forward pass through the simulator and one pass through the correction estimator. One important question is how our approach should treat different numbers of objects in the scene. When predicting physical parameter changes, the naive approach would be to predict all parameter updates at the same time. However, since the shape of the output is fixed at runtime, this would mean the network is only usable for a fixed number of objects. In our case, we decided to gain flexibility by predicting parameter changes for objects one at a time. We select the object of interest on the input by marking it with a different color, see \cref{fig2}. In the bin picking use case, this would amount to an initial segmentation task that can either be performed separately, or by the correction estimator itself. Comparing the single-object and multi-object approaches, no significant difference in performance was observed. 
An alternative approach to color-marking would be to place the objects themselves in separate channels, e.g. by instance segmentation.

\subsection{Simulator Module}
In principle, with our approach, the choice of the physics engine is arbitrary. For testing in a two-dimensional environment, we chose the Python-based physics library Pymunk\footnote{\url{http://www.pymunk.org}}, which is built on the physics engine Chipmunk\footnote{\url{http://chipmunk-physics.net}}.  The library allows for efficient simulation of two-dimensional rigid-body physics. Objects can be defined with arbitrary (two-dimensional) shapes, represented through polygons. Other physical properties like the mass, center-of-mass, moment of inertia, elasticity, and friction can be set as well, which allows for several different parameters to be predicted by our framework.  Forces can be added to objects, which allows for dynamical behavior. Fixed (immovable) objects can be added to the scene that can act as barriers. Pymunk includes helper functions for visualization with pygame\footnote{\url{https://www.pygame.org}}, which allows for prototyping and visualization of our test scenes.

\begin{figure}
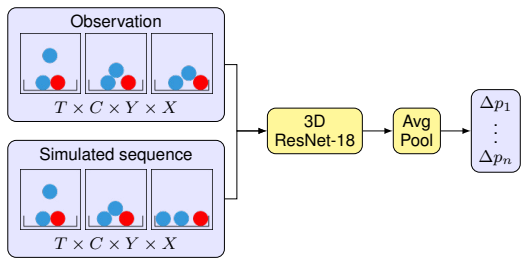

\centering
\adjustbox{scale=0.8}{%
\begin{tikzpicture}[font=\sffamily\footnotesize,
  m/.style={draw=black, rounded corners, fill=yellow!50, align=center},
  v/.style={m,fill=blue!10}
  ]

 \node[v,anchor=east,minimum width=3.6cm] (observed) {
    Observation\\[.1cm]
    
    \includegraphics[height=1cm]{figures/start_scene.png}
    \includegraphics[height=1cm]{figures/scene_2_1.png}
    \includegraphics[height=1cm]{figures/scene_2_2.png}\\[0cm]
    
    $T \times C \times Y \times X$
  };
 \node[v,below=0.3cm of observed,minimum width=3.6cm] (simseq) {
    Simulated sequence \\[.1cm]

    \includegraphics[height=1cm]{figures/start_scene.png}
    \includegraphics[height=1cm]{figures/scene_1_1.png}
    \includegraphics[height=1cm]{figures/scene_1_2.png}\\[.0cm]
    
    $T \times C \times Y \times X$
 };
 
 \coordinate (inbet) at ($(observed.south east)!0.5!(simseq.north east)$);
 
 \node[m,anchor=west] (net) at ($(inbet)+(0.7,0)$) {3D\\ResNet-18};
 
 \node[m,right=0.5cm of net] (pool) {Avg\\Pool};
 
 \node[v,right=0.5cm of pool] (out) {$\Delta p_1$\\$\vdots$\\$\Delta p_n$};
 
 \draw[-latex] (observed.east) -- ++(0.2,0) |- (net);
 \draw[-latex] (simseq.east) -- ++(0.2,0) |- (net);
 
 \draw[-latex] (net) -- (pool);

 \draw[-latex] (pool) -- (out);
 
\end{tikzpicture}}

\caption{Architecture of the learned correction estimator. The batch dimension is omitted.}
\label{fig:nn_arch}
\end{figure}

\subsection{Correction Estimator}
One possible choice for the correction estimator is to use generic gradient-free optimization techniques. In our case, we employ the \textit{hyperopt} library~\citep{hyperopt} for Python, which implements a Tree of Parzen Estimators~(TPE) approach to find the optimal set of parameters for a given objective function~\citep{bergstra2011}.  The objective function has to take the parameters as input and provide a measure for the accuracy of the output. As a criterion for accuracy, we choose the mean squared error between the ground truth and simulated time series in image-space.  Hyperopt logs arbitrary measures during the optimization cycle and provides the best set of parameters after optimization.
We expected that Hyperopt gives a reasonable baseline for optimization.

An alternative choice for the correction estimator is to use a neural network,
which learns the task of predicting optimal parameter updates.
The expectation is that a learned estimator can exploit the characteristics
of the underlying system much better than a generic optimizer.

Image-like representations and their time series are well-suited as an input representation, since they are readily available in our application domain (e.g. from semantic segmentation of the scene) and contain the necessary details, such as object position and contact information.
Therefore, our network directly operates on this representation.
The network architecture closely resembles a ResNet-18 architecture \cite{he2015}, however, we use 3D convolutional layers to accommodate for the time series collection of pictures (see~\cref{fig:nn_arch}). The input to our network has the shape $(B \times 2T \times C \times Y \times X)$, where $B$ is the batch size (10 or 20, with no discernable difference in performance), $2T$ the length of the concatenated time series (usually $2 \times 30 = 60$), $C$ the number of color / object channels and $Y \times X$ the resolution of the frames. We furthermore found that in our case, dropout regularization leads to more stable results as compared to the usually employed batch normalization. The network is trained to predict correction estimates between randomly chosen parameters ($p$ and $p^\prime$ for observation and guess time series, respectively) for a randomly chosen object in the scene.  The loss function is the mean squared error between the ground truth parameter correction $\Delta p_{\text{gt}} = p - p^\prime$, and guessed parameter correction by the correction estimator. In comparison to the Hyperopt approach, the network predicts a correction of the parameters that go into the simulator $\Delta p_i$ (as compared to the absolute value $p_i$).  The updated value can then be calculated by incrementing the old value with the parameter update $p_i^{(t+1)} = p_i^{(t)} + \Delta p_i^{(t)}$.

\begin{figure}
\centering
\includegraphics[width=\linewidth]{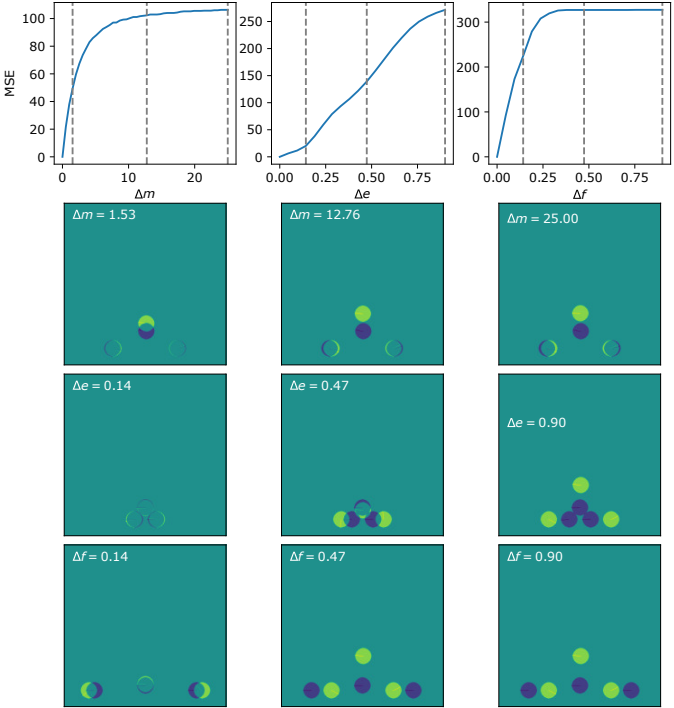}
\caption{Image-space MSE between two full time series with different physical parameters. The parameter under investigation is fixed to a specific value ($m_1=5$ and $f_1 = e_1 = 0.1$ for mass, friction, and elasticity) for one of the time series and varied for the other ($p_2 = p_1 + \Delta p$ for $p \in (m, e, f)$).
The pictures in the lower three rows show the difference between the last frame of each time series, respectively. The grey dashed lines in the top three plots indicate the values of the pictures in the bottom three rows (each panel in the first row from left to right represents a row from top to bottom). }
\label{fig4}
\end{figure}

\begin{figure*}
\centering
\includegraphics[scale=0.65,trim=0 10 0 10,clip]{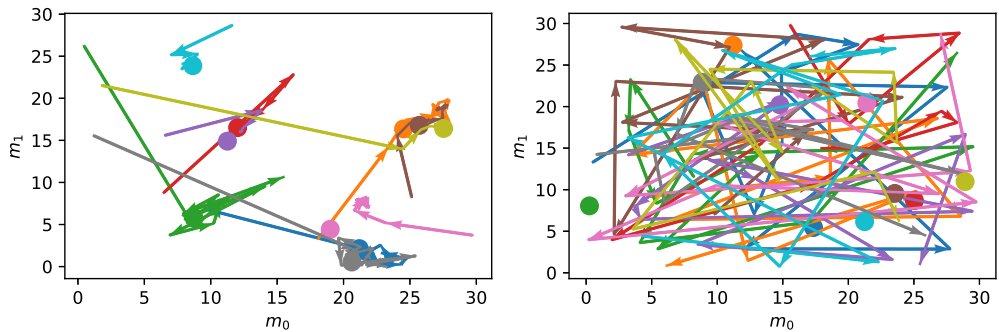}
\caption{Parameter space trajectories for two unknown masses $m_0$ and $m_1$ for the three circle scene (\cref{fig2}a). Shown are a neural network (left) and Hyperopt (right) as correction estimator. Each color represents a new scene instance, with optimization
iterations shown as arrows and the ground truth parameters as a filled circle.}
\label{fig5}
\end{figure*}

\begin{table*}
\centering
\begin{threeparttable}
\caption{Estimation errors}\label{table1}
\begin{tabular}{lrrrr}
\toprule
Configuration & Mass & Elasticity & Friction & Mass + Elasticity \\
\midrule
3 circles & $(2.6 \pm 0.9) \times 10^{-3}$ & $(8.7 \pm 1.9) \times 10^{-5}$ & - & $(1.4 \pm 0.5) \times 10^{-2}$   \\
3 circles (Hyperopt) & $(1.1 \pm 0.3) \times 10^{-2}$& - & - & $(5.2 \pm 0.7) \times 10^{-2}$  \\
second order & $(3 \pm 0.7) \times 10^{-4}$ & $(4 \pm 2) \times 10^{-3}$ & - & -    \\
box scene & $(2 \pm 0.5) \times 10^{-3}$ & - & $(3.9 \pm 1.2) \times 10^{-3}$& -    \\
bouncing balls & $(6.3 \pm 0.8) \times 10^{-3}$ & - & - & -  \\
\bottomrule
\end{tabular}
We show the normalized minimum achieved mean squared errors between predicted and true values after 11 iterations. Each mean is taken over 10 runs with random starting parameters. We also show standard deviations for each error.
\end{threeparttable}
\end{table*}

\subsection{Multiple Objects}
To keep our approach flexible with regards to the number of objects present in the scene, we highlight one of the objects by color.  The correction estimator then predicts parameter updates for the highlighted object. Alternative approaches are to place the investigated object in a different image channel, or fix the output dimension to allow for simultaneous prediction of several objects' properties. Both approaches have been tested and were found to yield similar performance.  While highlighting single objects requires $N_\text{obj}$ more forward passes through the prediction module, it allows for more flexibility.
We note that the required forward passes may be performed in parallel.

\section{Evaluation}

We evaluate our approach in two different scenarios with different combinations of predicted parameters and object configurations---a simple bin configuration and a more complex pool table setup. We compare the performance of Hyperopt and the neural network as correction estimator. Our measure of performance is the minimum achieved mean squared error between guessed and real parameters. To be able to compare different parameters with different number ranges, each parameter is normalized by its maximum achievable value. In case of unknown ground truth values, the minimum achieved error can be found by comparing the mean squared error between the time series, similar to the objective function of the Hyperopt approach.
Throughout our testing, runtime was dominated by physics simulation, not correction estimation. As a consequence, the number of iterations needed to find good parameters directly measures the performance of the approach for the correction estimator.

The learning-based estimator was trained using the Adam optimizer (learning rate \num{5e-6}) on 130000 simulated scenes. We note that it does not need to be trained on a specific scene configuration. The required training time can thus be done before deployment in a particular application.

\begin{figure}
\centering
\begin{tikzpicture}[font=\sffamily\footnotesize]
\node (pic) {\includegraphics[width=.8\linewidth]{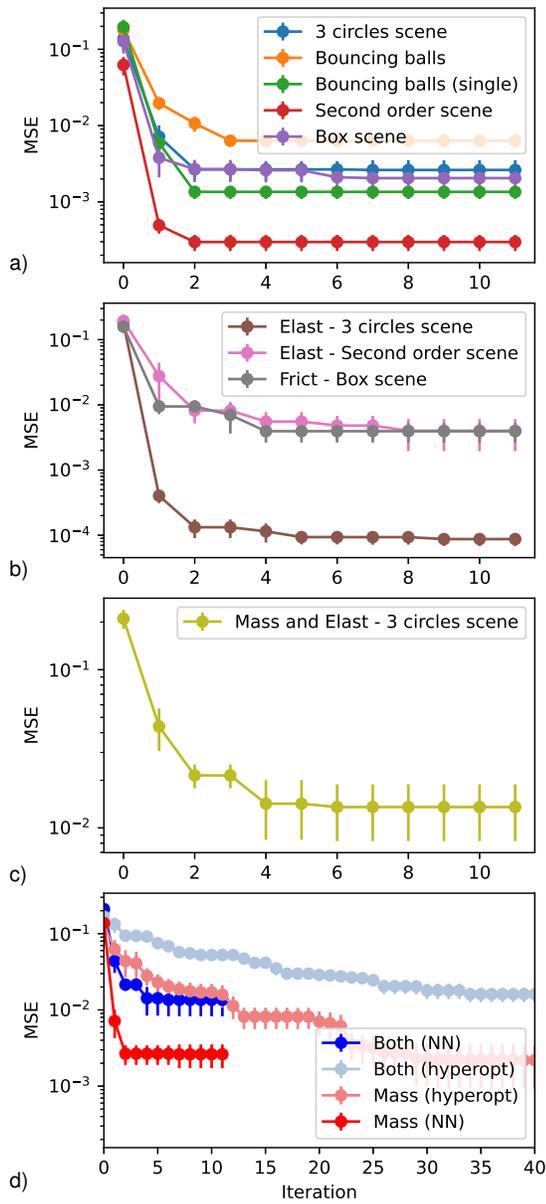}};
\node[anchor=south west,inner sep=0] at (rel cs:x=0,y=77,name=pic) {a)};
\node[anchor=south west,inner sep=0] at (rel cs:x=0,y=52,name=pic) {b)};
\node[anchor=south west,inner sep=0] at (rel cs:x=0,y=27,name=pic) {c)};
\node[anchor=south west,inner sep=0] at (rel cs:x=0,y=2,name=pic) {d)};
\end{tikzpicture}
\caption{Minimum normalized Mean Squared Error between predicted and true parameters for different scene and parameter combinations: a) mass, b) elasticity \& friction, c) mass \& elasticity.
d) shows a comparison between hyperopt and the learned network for mass and mass \& elasticity.
Note that we show the best parameters until each iteration, with the
image-space MSE used to determine the so-far-found optimum.}
\label{fig3}
\end{figure}

\subsection{Objects in a Bin}
The first investigated scene is inspired by bin picking scenarios. It consists of a two-dimensional box formed by immovable lines, with three objects placed in it. One of the observed objects is a test object with constant physical parameters throughout all scene iterations. This object fixes the numerical value for the other objects, as the collisions dynamics between objects only depends on the ratio between their physical parameters. All objects are subject to gravitational force, and the test object is additionally accelerated towards the other objects. In this scenario, we investigate three different scenes: 1) Three circles, where the test object is accelerated towards the two unknown object from above, 2) second-order collisions, where the test object is accelerated from one direction towards one of the unknown objects, which then interacts with the other unknown object and 3) stacked boxes, with two square-shaped objects stacked on top of each other, where the circular shaped test object is accelerated towards the stacked objects. These scenes are depicted in \cref{fig2}~(a) to (c).

To verify predictiveness of the different physical parameters, we investigate how the mean squared error between the raw time series changes for different values of the predicted parameters, see \cref{fig4}, topmost row. This analysis shows, that for both mass and elasticity, a gradient towards small parameter changes always exists (albeit smaller for mass). For friction on the other hand, the curve looks flat above a certain threshold.  This can be explained by physical considerations: To first order the collisional dynamics of rigid circles does not depend on the surface friction, therefore, changing the friction does not influence the trajectory of the objects. \cref{fig4} also shows the difference between the final frames of the compared timeseries for different parameter values. This highlights the necessity for expressive dynamics, that yields information about the physical parameters under investigation.

We train the network on each scene separately on randomly chosen parameters to predict parameter changes. To evaluate the performance, we generate new scenes with random parameters and predict parameter updates for each mass starting with a random guess. We then resimulate and iterate this procedure.  We find that our approach generally converges towards its final value after one or two iterations, see \cref{fig3}. For comparison, we test generic hyperparameter optimization to predict the parameters, which finds comparably good values after an order of magnitude more iterations, see \cref{fig3}~(bottom).  This is also highlighted in parameter space, where the Hyperopt approach performs much more exploration, see \cref{fig5}. The best value after 11 iterations for each tested scene / parameter combination is shown in \cref{table1}. In agreement with the results of \cref{fig4}, we found that for some combinations of scenes and parameters, the performance of the NN drops drastically, for example when predicting the friction in scene 1). However, in a more expressive scene with regard to friction, i.e. scene 3), we find comparable performance. From \cref{fig4}, we also find that the image-space MSE has a favorable shape for the elasticity in the three circle scene, which is reflected in the superior performance when predicting elasticity.

We also investigate the question whether the network can handle scene observations of different nature than the sharp segmentations provided by the simulator. For this purpose, we apply a gaussian filter with $\sigma=3$\,pixels to the observations,
so that we obtain a more fuzzy observation. The simulated scenes are, however, supplied in their original, sharp version. In our experiments we did not observe significant changes in the performance of the neural network as a correction estimator.

\subsection{Bouncing Balls}
The second scenario investigates the role of a variable number of objects in the scene.  We use the well-known "bouncing balls" scene (a two-dimensional representation of a "pool table"), see \cref{fig2}d. We place one test object with fixed mass on a pool table that is bounded by four rigid lines. The test mass is accelerated towards a variable number of balls with unknown mass that are placed in a triangular pattern, such that all balls collide with at least one other object. We predict each objects mass separately by marking it with a different color. We randomly place between two and six balls with random mass on the table.  To evaluate the performance of our approach,  we update each objects mass as predicted by the network and then iterate over the updated guesses.  Even in this more complex state space, we find only slightly reduced accuracy after several iterations. However, in comparison to fewer objects, it takes about three to four iterations for the values to converge. We furthermore find that the network is able to predict a scene with just one unknown object with high accuracy, although this setup was not part of the training process, see \cref{fig3}a, green line.

\section{Limitations}

Our method makes several assumptions, which will be discussed here. For instance,
the system is currently limited to 2D scenes, but this is not an inherent
constraint. Furthermore, we assume that the observed state can be easily compared
to the simulated one, in our case by rendering a similar view from the simulation
data. As (visual) scene complexity increases, this assumption may not hold anymore.

\section{Conclusion}
In conclusion, we find that our approach of predicting physical object properties from video allows for much more rapid refinement of parameters in comparison to generic gradient-free parameter optimization techniques.  This allows for a fast determination of physical parameters from a video input. The accuracy on all parameters is $10 \%$ or better. For higher accuracy, it may be necessary to either finetune the used neural network, or use a generic optimization approach or a differentiable physics engine as a second step in the optimization process.  Due to the iterative nature and fast convergence of our approach, it allows for online refinement of physical parameters. In any case, the accuracy of the estimation strongly depends on the expressiveness of the scene with regard to the investigated parameters.

\section*{Acknowledgments}

This work was funded by grant BE 2556/18-2 (Research Unit FOR 2535
Anticipating Human Behavior) of the German Research Foundation (DFG).

\printbibliography

\end{document}